# An Evolving Cascade System Based on A Set Of Neo-Fuzzy Nodes

**Zhengbing Hu**
School of Educational Information Technology, Central China Normal University, Wuhan, China
Email: hzb@mail.ccnu.edu.cn

**Yevgeniy V. Bodyanskiy**
Kharkiv National University of Radio Electronics, Kharkiv, Ukraine,
Email: yevgeniy.bodyanskiy@nure.ua

**Oleksii K. Tyshchenko and Olena O. Boiko**
Kharkiv National University of Radio Electronics, Kharkiv, Ukraine,
Email: lehatish@gmail.com, olena.boiko@ukr.net

*Abstract*— Neo-fuzzy elements are used as nodes for an evolving cascade system. The proposed system can tune both its parameters and architecture in an online mode. It can be used for solving a wide range of Data Mining tasks (namely time series forecasting). The evolving cascade system with neo-fuzzy nodes can process rather large data sets with high speed and effectiveness.

## I. Introduction

The task of time series forecasting (data sequences forecasting) is well studied nowadays. There are many mathematical methods of different complexity that can be used for solving this task: spectral analysis, exponential smoothing, regression, advanced intellectual systems, etc. In many real-world cases, analyzed time series are non-stationary, nonlinear, and usually contain unknown behavior trends, stochastic or chaotic components. This obstacle complicates time series forecasting and makes the above mentioned systems less effective.

To solve this problem, nonlinear models based on mathematical methods of Computational Intelligence [1-3] can be used. It should be especially mentioned that neuro-fuzzy systems [4-6] are widely used for this type of tasks due to their approximating and extrapolating properties, results' interpretability, and learning abilities. The most appropriate choice for non-stationary data processing is evolving connectionist systems [7-10]. These systems adjust not only their synaptic weights and parameters of membership functions, but also their architectures.

There are many evolving systems that are able to process data sets in an online mode. Most of them are based on multilayer neuro-fuzzy systems. The Takagi-Sugeno-Kang (TSK) fuzzy systems [11-12] and adaptive neuro-fuzzy inference systems (ANFIS) are the most popular and effective systems that are used to solve such tasks. But in some cases (e.g. when a size of a data set is not sufficient for training) they cannot rapidly tune their parameters, so their effectiveness can decrease.

The first solution for this problem is to decompose an initial task into a set of simpler tasks, so that the obtained system can solve a problem with a data set at hand regardless to its size. One of the most studied approaches based on this principle is the Group Method of Data Handling (GMDH) [13-14]. But in case of online data processing, the GMDH systems are not sufficiently effective. This problem can be solved by an evolving cascade model that tunes both its parameters as well as its architecture in an online mode.

Generally speaking, one can use different types of neurons or other more complicated systems as nodes in an evolving cascade system. For example, a compartmental R-neuron was introduced as a node of a cascade system [15, 16]. If a data set to be processed is large enough, it seems reasonable to use neo-fuzzy neurons [17-19]. The neo-fuzzy neuron is capable of finding a global minimum for a learning criterion in an online mode, it also has a high learning speed and good approximating properties. It is also appropriate from the viewpoint of computational simplicity.

The remainder of this paper is organized as follows: Section 2 describes an architecture of the evolving cascade system. Section 3 describes an architecture of the neo-fuzzy neuron as a node of the evolving cascade system. Section 4 presents several synthetic and real-world applications to be solved with the help of the proposed evolving cascade system. Conclusions and future work are given in the final section.

## II. AN EVOLVING CASCADE MODEL

An architecture of the evolving cascade model is shown in Fig. 1.

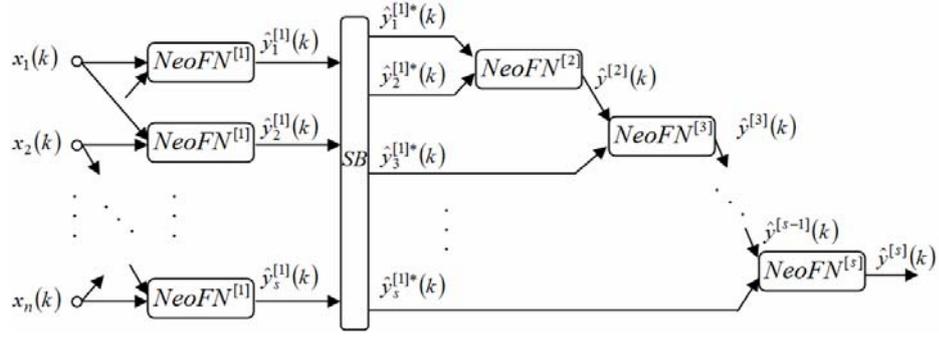

Figure 1. An architecture of the evolving cascade model

A $(n \times 1)$-dimensional vector of input signals $x(k) = \left( x_1(k), x_2(k), ..., x_n(k) \right)^{\mathrm{T}}$ is fed to the zero layer of the system (here $k = 1, 2, ...$ is an index of the current discrete time). The first hidden layer contains $c_n^2$ nodes (each one has two inputs). The outputs of the first hidden layer form signals $\hat{y}_s^{[1]}(k)$, $s = 1, 2, ..., 0,5n(n-1) = c_n^2$. Then these signals are fed to the selection block $SB$ that sorts the first layer nodes by some criteria (e.g. by the mean squared error $\sigma_{\hat{y}_s^{[1]}(k)}^2$) so that $\sigma_{\hat{y}_1^{[1]*}(k)}^2 < \sigma_{\hat{y}_2^{[1]*}(k)}^2 < ... < \sigma_{\hat{y}_s^{[1]*}(k)}^2$.

The outputs $\hat{y}_1^{[1]}*(k)$ and $\hat{y}_2^{[1]}*(k)$ of the selection block are fed to a unique neuron of the second layer which forms its output signal $\hat{y}^{[2]}(k)$. This signal and the signal $\hat{y}_3^{[1]}*(k)$ (an output signal of the selection block $SB$) are fed to a node of the next layer. A process of the cascades' increasing is continued until a required accuracy is obtained.

## III. THE NEO-FUZZY SYSTEM AS A NODE OF THE EVOLVING CASCADE SYSTEM

Neo-fuzzy neurons were proposed by T. Yamakawa and co-authors [17-19]. Advantages of this block are good approximating properties, computational simplicity, a high learning speed, and ability of finding a global minimum for a learning criterion in an online mode. An architecture of the neo-fuzzy neuron as a node of the evolving cascade system is shown in Fig.2.

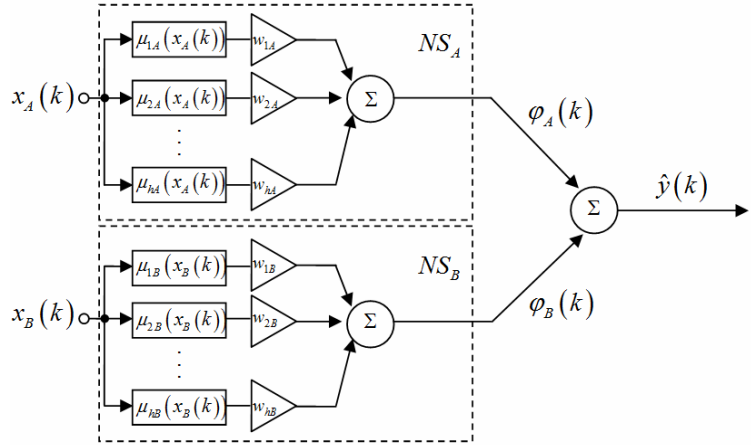

Figure 2. An architecture of the neo-fuzzy neuron

Nonlinear synapses $NS_A$ and $NS_B$ which are structural elements of the neo-fuzzy neuron fulfill the Takagi–Sugeno fuzzy inference of the zero order. A two-dimensional vector of input signals $x(k) = \left( x_A(k), x_B(k) \right)^{\mathrm{T}}$ is fed to the node's input. The first layer of each nonlinear synapse contains $h$ membership functions. In [20], it was proposed to use the B-splines as membership functions for the neo-fuzzy neuron. B-splines provide higher approximation quality. A B-spline of the $q$-th order has the form

$$\mu_{iA}^{q}\left(x_A\left(k\right)\right)=\begin{cases}\left.\begin{array}{l}1\text{ if }c_{iA}\leq x_A\left(k\right)<c_{i+1,A},\\0\text{ otherwise}\end{array}\right\}\text{ for }q=1\\[2em]\dfrac{x_A\left(k\right)-c_{iA}}{c_{i+q-1,A}-c_{iA}}\mu_{iA}^{q-1}\left(x_A\left(k\right)\right)+\\+\dfrac{c_{i+q,A}-x_A\left(k\right)}{c_{i+q,A}-c_{i+1,A}}\mu_{i+1,A}^{q-1}\left(x_A\left(k\right)\right)\text{ for }q>1\\[2em]i=1,...,h-q,\end{cases}$$

$$\mu_{iB}^{q}\left(x_B\left(k\right)\right)=\begin{cases}\left.\begin{array}{l}1\text{ if }c_{iB}\leq x_B\left(k\right)<c_{i+1,B},\\0\text{ otherwise}\end{array}\right\}\text{ for }q=1\\[2em]\dfrac{x_B\left(k\right)-c_{iB}}{c_{i+q-1,B}-c_{iB}}\mu_{iB}^{q-1}\left(x\left(k\right)\right)+\\\dfrac{c_{i+q,B}-x_B\left(k\right)}{c_{i+q,B}-c_{i+1,B}}\mu_{i+1,B}^{q-1}\left(x_B\left(k\right)\right)\text{ for }q>1\\[2em]i=1,...,h-q.\end{cases}$$

where $c_{iA}$, $c_{iB}$ are parameters that define centers of the membership functions. It should be noticed that when $q=2$ one can get traditional triangular membership functions, and when $q=4$ one can get cubic splines, etc. The B-splines meet the unity partitioning conditions

$$\begin{cases}\sum_{p=1}^{h}\mu_{pA}\left(x_A\left(k\right)\right)=1,\\\sum_{p=1}^{h}\mu_{pB}\left(x_B\left(k\right)\right)=1\end{cases}$$

that allows to simplify the node's architecture excluding a normalization layer.

So, the elements of the first layer compute membership levels $\mu_{pA}\left(x_A\left(k\right)\right)$, $\mu_{pB}\left(x_B\left(k\right)\right)$, $p=1,2,...,h$.

The second layer contains synaptic weights $w_{iA}$, $w_{iB}$ that are adjusted during a learning process.

The third layer is formed by two summation units. It computes sums of the output signals of the second layer for each nonlinear synapse $NS_A$ and $NS_B$. The outputs of the third layer are values

$$\begin{cases}\varphi_A\left(k\right)=\sum_{p=1}^{h}w_{pA}\mu_{pA}\left(x_A\left(k\right)\right),\\\varphi_B\left(k\right)=\sum_{p=1}^{h}w_{pB}\mu_{pB}\left(x_B\left(k\right)\right).\end{cases}$$

Another summation unit sums up these two signals in order to produce the output signal $\hat{y}\left(k\right)$ of the node

$$\hat{y}(k) = \varphi_A(k) + \varphi_B(k) =$$
$$= \sum_{p=1}^{h} w_{pA} \mu_{pA}(x_A(k)) + \sum_{p=1}^{h} w_{pB} \mu_{pB}(x_B(k)) \quad (1)$$

The expression (1) can be written in the form

$$\hat{y}(k) = w^T(k)\varphi(k)$$

where
$$w(k) = (w_{1A}(k), ..., w_{hA}(k), w_{1B}(k), ..., w_{hB}(k))^T,$$

$$\varphi(k) = (\mu_{1A}(x_A(k)), ..., \mu_{hA}(x_A(k)), \mu_{1B}(x_B(k)), ..., \mu_{hB}(x_B(k)))^T.$$

To learn the neo-fuzzy neuron, we can use the procedure [21, 22]

$$\begin{cases} w(k) = w(k-1) + r^{-1}(k)\left(y(k) - w^T(k-1)\varphi(k)\right)\varphi(k) \\ r(k) = \alpha r(k-1) + \varphi^T(k)\varphi(k), 0 \le \alpha \le 1 \end{cases} \quad (2)$$

which possesses both filtering and tracking properties. It can be noticed that when $\alpha = 1$ the procedure (2) coincides with the Kaczmarz–Widrow–Hoff optimal gradient algorithm

$$w(k) = w(k-1) + \frac{y(k) - w^T(k-1)\varphi(k)}{\varphi^T(k)\varphi(k)}\varphi(k)$$

that can be used if a training data set is non-stationary [23].

## IV. EXPERIMENTS

In order to prove the effectiveness of the proposed system, several simulation tests were implemented. The system's effectiveness was analyzed by a value of the root mean square error (RMSE) and data processing time.

### A. A nonlinear system

A nonlinear system to be identified can be described by the equation

$$y_t = \frac{\sum_{i=1}^{m} y_{t-i}}{1 + \sum_{i=1}^{m} (y_{t-i})^2} + u_{t-1}$$

where $u_t = \sin(2\pi t / 20)$, $y_j = 0$, $j = 1, ..., m$, $m = 10$. The model is presented in the form

$$\hat{y}_t = f(y_{t-1}, y_{t-2}, ..., y_{t-10}, u_{t-1})$$

where $\hat{y}_t$ stands for a model's output. The aim was to predict the next output using previous inputs and outputs.

This data set contains 2500 points: 2000 points were selected for a training stage and 500 points were used for testing.

To compare results, we used a multilayer perceptron (MLP), a radial-basis function neural network (RBFN), an adaptive neuro-fuzzy inference system (ANFIS), and the proposed evolving cascade system with neo-fuzzy nodes.

MLP was learnt during five epochs. A number of MLP's inputs was equal to 5 and a number of hidden nodes was equal to 10. A total number of parameters to be tuned was equal to 51.

In the RBFN's architecture, we used 3 inputs and 11 kernel functions, so a total number of parameters to be tuned in the RBFN's system was roughly equal to 56.

One of the best results was shown by ANFIS, but it was processing data during 5 epochs and a little longer than the proposed system, so it cannot process data in an online mode. ANFIS had 3 inputs and 34 hidden nodes. A total number of parameters to be adjusted was equal to 32.

The proposed evolving cascade system had 3 inputs and 4 membership functions in each node. A total number of parameters to be adjusted was 54. The proposed system demonstrated the best prediction results, and its data processing time was the best among the others.

A comparison of the systems' results is shown in Table 1.

TABLE I. COMPARISON OF THE SYSTEMS' RESULTS

| Systems | Parameters to be tuned | RMSE (training) | RMSE (test) | Time, s |
|---------|------------------------|-----------------|-------------|---------|
| MLP | 51 | 0.0173 | 0.0178 | 0.5313 |
| RBFN | 56 | 0.0990 | 0.0993 | 0.4828 |
| ANFIS | 32 | 0.0070 | 0.0085 | 0.3625 |
| The proposed system | 54 | 0.0413 | 0.0462 | 0.3281 |

Prediction results for the proposed system are in Fig.3 (a blue line represents signal's values, a magenta line represents prediction values, and a grey line represents prediction errors).

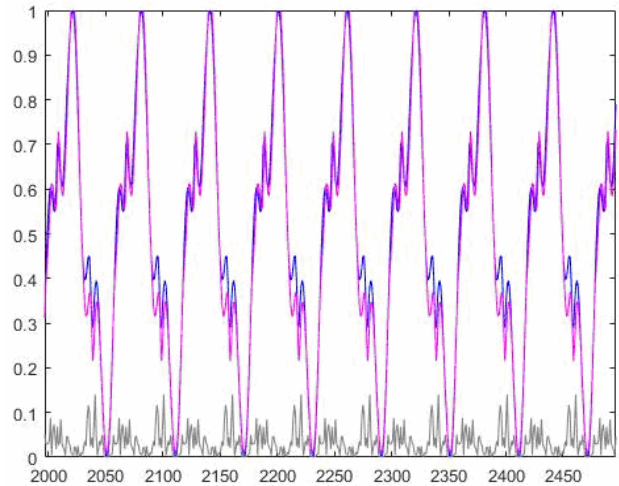

Figure 3. Prediction results

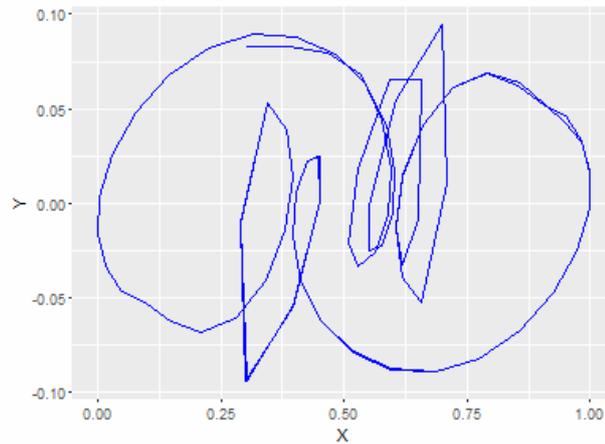

Figure 4. A phase portrait of the signal

### B. Internet traffic data (in bits) from an ISP

This data set describes hourly traffic in the United Kingdom academic network backbone (taken from datamarket.com). It was collected between November 19[th], 2004, at 09:30 and January 27[th], 2005, at 11:11. This data set contains 1657 points: 1326 points were selected for a training stage and 331 points were used for testing.

To compare results, a similar to the previous experiment set of systems was used but with other characteristics.

MLP was learnt during one epoch (this case is similar to learning in an online mode). A number of MLP's inputs was equal to 3 and a number of hidden nodes was equal to 8. A total number of parameters to be tuned was 31.

In the RBFN's architecture, 7 kernel functions and 3 inputs were used, so a total number of parameters to be adjusted in the RBFN system was roughly equal to that of the proposed system.

ANFIS had 4 inputs and 55 hidden nodes. It was processing data during 5 epochs. A total number of parameters to be tuned was 80. This system showed the best prediction result, but it processed data longer than the proposed system.

The proposed evolving cascade system had 4 inputs and 4 membership functions in each node. A total number of parameters to be tuned was 36. The proposed system showed one of the best prediction results according to RMSE (the second result after ANFIS), and its data processing time was the best among the others.

A comparison of the systems' results is shown in Table 2.

TABLE II. COMPARISON OF THE SYSTEMS' RESULTS

| Systems | Parameters to be tuned | RMSE (training) | RMSE (test) | Time, s |
|---|---|---|---|---|
| MLP | 31 | 0.0682 | 0.0755 | 0.2656 |
| RBFN | 36 | 0.1038 | 0.1114 | 0.2562 |
| ANFIS | 80 | 0.0265 | 0.0270 | 0.2031 |
| Proposed system | 36 | 0.0636 | 0.0550 | 0.1718 |

Prediction results for the proposed system are in Fig.5.

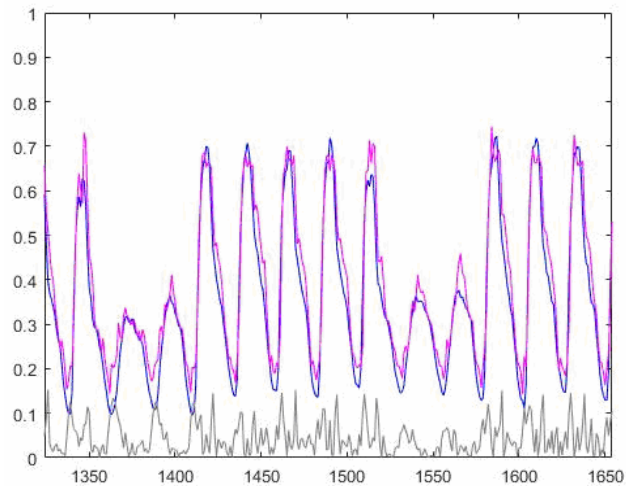

Figure 5. Prediction results

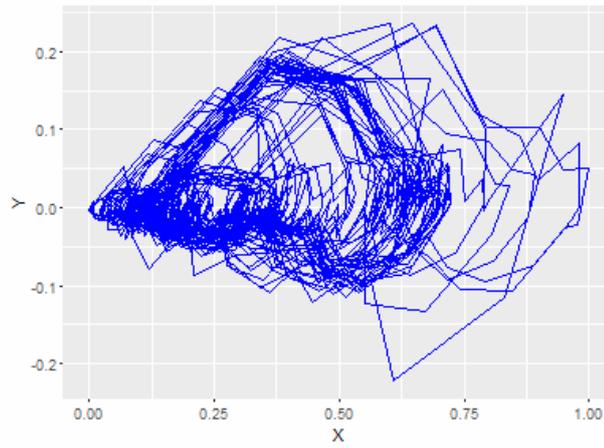

Figure 6. A phase portrait of the signal

### C. Darwin Sea Level Pressures

This data set was taken from research.ics.aalto.fi. This data set describes monthly values of the Darwin sea level pressure. It was collected between 1882 and 1998. This data set contains 1300 points: 1040 points were selected for a training stage and 260 points were used for testing.

MLP was learnt during five epochs. It had 3 inputs and 10 hidden nodes. A total number of parameters to be adjusted was equal to 41.

RBFN had 7 kernel functions and 3 inputs, and a total number of parameters to be tuned in the RBFN system was equal to 36, i.e. it was very close to a number of parameters in the proposed system.

A number of inputs for ANFIS was 4, a number of hidden nodes was equal to 55. ANFIS was processing data during one epoch. A total number of parameters to be tuned was equal to 80. This system showed the best result from the view point of accuracy, but it processed data longer than the proposed system.

The proposed evolving cascade system had 3 inputs and 4 membership functions in each node. A total number of parameters to be tuned was 40. The proposed system showed one of the best results, and its data processing time was the best.

A comparison of the systems' results is shown in Table 3.

TABLE III. COMPARISON OF THE SYSTEMS' RESULTS

| Systems | Parameters to be tuned | RMSE (training) | RMSE (test) | Time, s |
|---|---|---|---|---|
| MLP | 41 | 0.0843 | 0.0886 | 0.4844 |
| RBFN | 36 | 0.1495 | 0.1512 | 0.2391 |
| ANFIS | 80 | 0.0756 | 0.0866 | 0.2031 |
| Proposed system | 40 | 0.1159 | 0.1483 | 0.1250 |

Prediction results for the proposed system are in Fig.7.

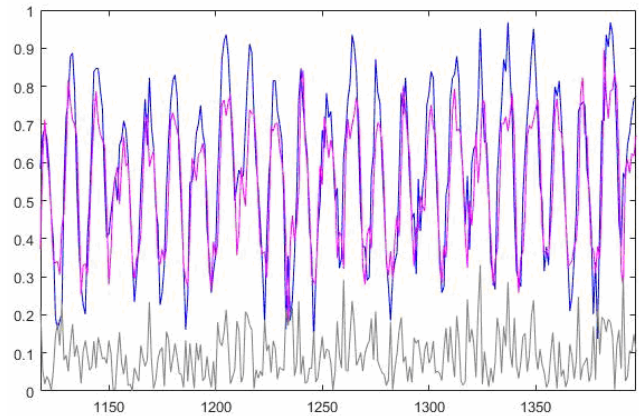

Figure 7. Prediction results

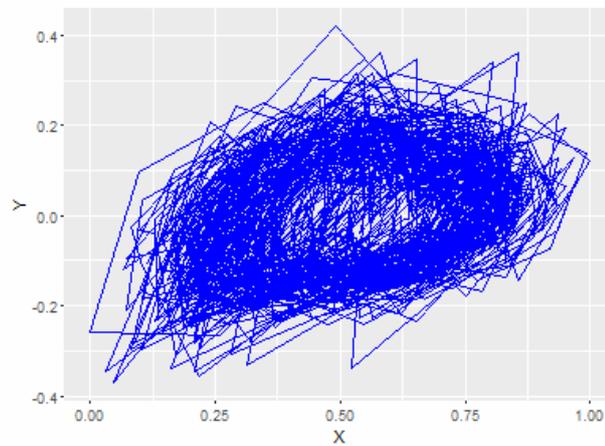

Figure 8. A phase portrait of the signal

## V. CONCLUSION

An evolving cascade model based on the neo-fuzzy nodes is proposed. It can adjust both its architecture in an online mode. The proposed system has a rather simple computational implementation and can process data sets with a high speed. A number of experiments demonstrated that this evolving cascade system can forecast time series with high effectiveness. The results may be successfully used in a wide class of Data Stream Mining and Dynamic Data Mining tasks.

## ACKNOWLEDGMENT

The authors would like to thank anonymous reviewers for their careful reading of this paper and for their helpful comments.

This scientific work was supported by RAMECS and CCNU16A02015.